# MR-EIT: Multi-Resolution Reconstruction for Electrical Impedance Tomography via Data-Driven and Unsupervised Dual-Mode Neural Networks

Fangming Shi, Jinzhen Liu, Xiangqian Meng, Yapeng Zhou, Hui Xiong,

*Abstract*—This paper presents a multi-resolution reconstruction method for Electrical Impedance Tomography (EIT), referred to as MR-EIT, which is capable of operating in both supervised and unsupervised learning modes. MR-EIT integrates an ordered feature extraction module and an unordered coordinate feature expression module. The former achieves the mapping from voltage to two-dimensional conductivity features through pre-training, while the latter realizes multi-resolution reconstruction independent of the order and size of the input sequence by utilizing symmetric functions and local feature extraction mechanisms. In the data-driven mode, MR-EIT reconstructs high-resolution images from low-resolution data of finite element meshes through two stages of pre-training and joint training, and demonstrates excellent performance in simulation experiments. In the unsupervised learning mode, MR-EIT does not require pre-training data and performs iterative optimization solely based on measured voltages to rapidly achieve image reconstruction from low to high resolution. It shows robustness to noise and efficient super-resolution reconstruction capabilities in both simulation and real water tank experiments. Experimental results indicate that MR-EIT outperforms the comparison methods in terms of Structural Similarity (SSIM) and Relative Image Error (RIE), especially in the unsupervised learning mode, where it can significantly reduce the number of iterations and improve image reconstruction quality.

## I. INTRODUCTION

Electrical Impedance Tomography (EIT) is an imaging technique based on electrical principles, which has significant advantages such as low cost and non-invasiveness. Its core principle is to measure the voltage distribution on the boundary of the region under test to infer the distribution of electrical conductivity within that region. After decades of development, EIT technology has been widely applied in various fields [1,2,3,4,5,6,7], including stroke imaging, breast cancer detection, and lung imaging. However, EIT image reconstruction faces many challenges. First, the propagation path of current within an object is affected by the distribution of electrical conductivity, making the image reconstruction problem highly nonlinear. This complicates the reconstruction process and increases computational difficulty. To achieve EIT image reconstruction, multiple electrodes are typically installed around the imaging domain. By applying a low-frequency current between a pair of electrodes and measuring the voltage between other pairs of electrodes, data is obtained. Using these measured voltage data, combined with complex mathematical models and algorithms, the distribution of electrical conductivity in the field under test is ultimately derived to complete the image reconstruction.

Recently, many methods have employed neural networks to achieve EIT image reconstruction. Zhang et al. [8] proposed a novel dense attention network for the precise reconstruction of lung contours and lesion structures. The DA-Net combines the dense connections of DenseNet with attention mechanisms, enhancing information flow and improving reconstruction accuracy. Huang et al. [9] developed a 2D EIT image reconstruction method based on a hybrid precision asymmetric neural network. Ma et al. [10] introduced a model-driven deep learning reconstruction network for EIT tactile sensing, named PDCISTA-Net. This network integrates a preprocessing filtering module with a dual-channel iterative shrinkage-thresholding algorithm (ISTA), capable of capturing the block correlation and sparsity in impedance change distribution. Sun et al. [11] proposed a multimodal EIT image reconstruction algorithm utilizing a deep similarity prior. They designed a self-supervised network comprising two structure encoding networks with shared weights and constructed multimodal EIT reconstruction through neural network regularization.

In the field of 3D reconstruction, Neural Radiance Fields (NeRF) has been widely used to generate realistic novel views of spatial scenes. NeRF synthesizes new views of complex scenes by optimizing a continuous volumetric scene function, which takes spatial positions and viewing directions as inputs and outputs the volume density and view-dependent emitted radiance at those positions. This method leverages classical volume rendering techniques to project the output colors and

This work was supported in part by the National Natural Science Foundation of China under Grant 62071329, in part by the Natural Science Foundation Applying System of Tianjin under Grant 18JCYBJC90400 and Grant 18JCQNJC84000, and in part by the Science and Technology Development Fund of Tianjin Education Commission for Higher Education under Grant 2019KJ014.

Fangming Shi is with the Key Laboratory of Intelligent Control of Electrical Equipment, Tiangong University, Tianjin 300387, China

Jinzhen Liu is with the School of Control Science and Engineering, Tiangong University, Tianjin 300387, China.

Xiangqian Meng is with the Key Laboratory of Intelligent Control of Electrical Equipment, Tiangong University, Tianjin 300387, China.

Hui Xiong is with the School of Artificial Intelligence, Tiangong University, Tianjin 300387, China (e-mail: xionghui@tiangong.edu.cn).

Yapeng Zhou is with the Key Laboratory of Intelligent Control of Electrical Equipment, Tiangong University, Tianjin 300387, China.

densities into images, thereby synthesizing high-quality novel views from sparse input views. This capability of continuous spatial representation can also be applied to image super-resolution tasks [12,13,14], treating images as continuous functions rather than discrete pixels. In the fields of image generation and 3D reconstruction, there has never been a clear boundary between segmentation and generative models. Effective methods for pixel or point cloud segmentation can generally be adapted with minor modifications to perform high-level generative tasks. For example, the classic structure of U-Net [15] can be used to form the generative network of diffusion models [16]. Similarly, the classic structure for point cloud segmentation can also be used to achieve reconstruction tasks. In the field of medical imaging, various implicit neural representation methods have already been applied to CT and MRI images [17,18].

Most EIT image reconstruction using deep learning techniques employs neural networks to establish an end-to-end mapping from voltage to conductivity. Since neural network designs are generally tailored for specific numbers of elements or pixels, it is inconvenient to reconstruct images of different resolutions. It is also difficult to quickly adapt when changing the shape of the field or using the same reconstruction field with a different number of mesh elements. When neural networks use operators that are independent of the sequence length, they can achieve reconstruction for various numbers of elements, but they cannot utilize local feature extraction methods like convolution. To enable the reconstruction algorithm to conveniently handle EIT image reconstruction at multiple resolutions and to benefit from excellent feature extraction due to local features under sequence order - independent conditions, we propose the MR - EIT method. Inspired by the image super - resolution method LIIF [14] and the classic PointNet series [19,20,21] in the field of point cloud segmentation, this method covers two ways of using neural networks to reconstruct EIT images, namely data - driven image reconstruction and unsupervised learning - based image reconstruction. Under data -driven conditions, an ordered feature extraction module suitable for voltage feature extraction is designed to achieve multi - grid - element image reconstruction through conditional implicit neural representation. Under unsupervised learning conditions, MR - EIT no longer requires training data; it only needs measured voltages to minimize errors for image reconstruction, and the unordered coordinate feature expression module serves as an implicit regularization.

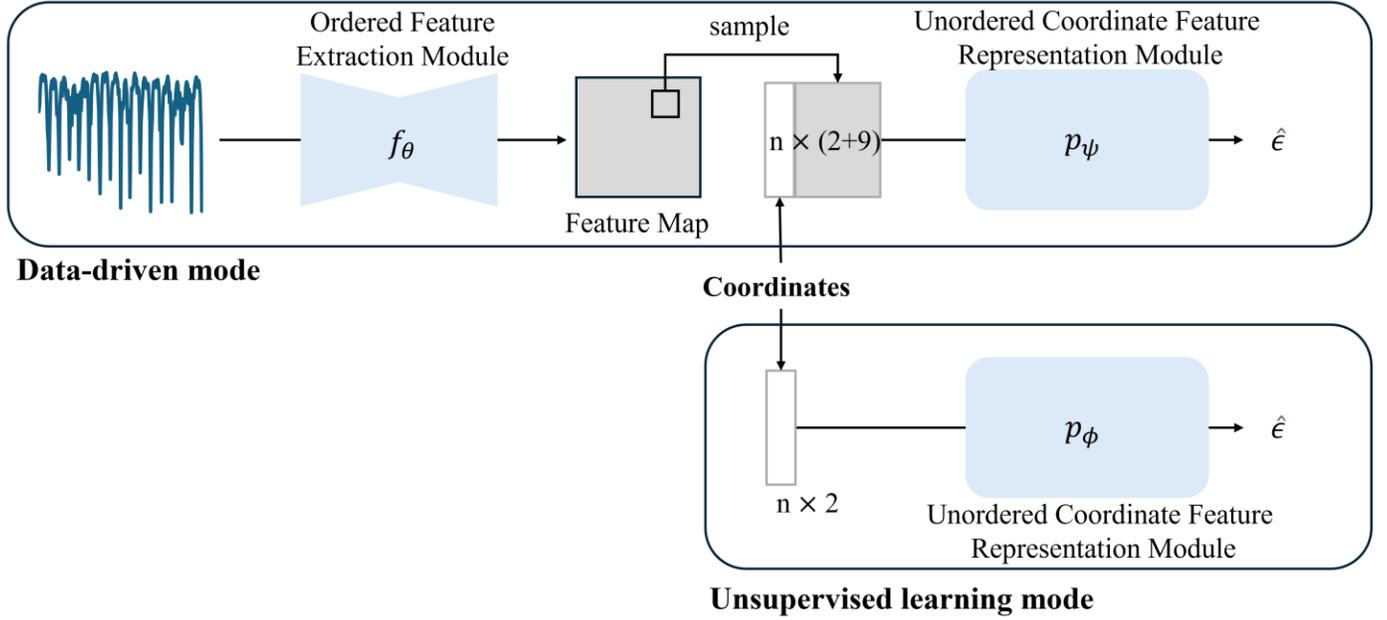

Fig. 1 The overall structure of MR-EIT

The main contributions of this paper are as follows:
- We propose the MR - EIT method, which can reconstruct images of various resolutions.
- Constructed a neural network architecture suitable for both data - driven and unsupervised learning modes.
- The performance of the algorithm under both modes was verified through simulations and physical experiments, especially under the unsupervised mode, where high - level image reconstruction can be achieved without pre - training the neural network.

Section II of this paper introduces the related work. Section III elaborates on the algorithm principles and neural network architectures of the two modes of MR - EIT. Section IV validates the image reconstruction performance of the model in both data - driven and unsupervised learning modes using simulated data and real - world experiments.

## II. RELATED WORK

### A. Local Implicit Image Function

The Local Implicit Image Function (LIIF) is a method for continuous image representation [14], designed to address the limitations of traditional pixel - based image representations in resolution conversion. LIIF encodes an image as a set of latent codes distributed in the spatial dimension and uses a shared decoding function to predict the RGB values at given coordinates, thereby enabling image rendering at arbitrary resolutions. Feature unfolding is employed to construct the feature vector of a pixel by combining the features of neighboring pixels when sampling the features of a specific coordinate:

$$\widehat{M}_{ij} = Concat(\{M_{i+l,j+m}\}_{l,m \in \{-1,0,1\}}) \quad (1)$$

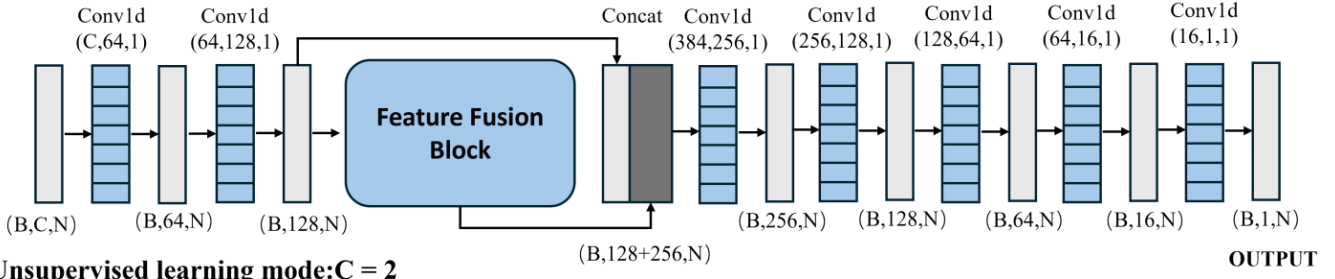

Fig. 2 Unordered Coordinate Feature Expression Module.

Here, $M$ represents the voxelized output latent encoding of LIIF, and $\widehat{M}$ is the feature vector that combines the features of the 3×3 neighboring pixels centered on the sampling coordinate.

### B. Point Cloud Segmentation

PointNet [19] and PointNet++ [20] are classic models for point cloud segmentation tasks, where max - pooling is used to extract global or local features. This mechanism based on symmetric function design has been proven to achieve SOTA performance compared with recent works [21]. For an unordered set of spatial coordinates, without voxelization, local feature encoding can be obtained by collecting the features of the nearest coordinates within a specific range of a coordinate $x^n, n \in R^d$ as $\{F_1, F_2 ..., F_n\}$ and using max – pooling:

$$\widehat{F_{x^n}} = \underset{i=1,\dots,n}{MAX}\{F_1, F_2 ..., F_n\} \quad (2)$$

where $\widehat{F_{x^n}}$ is the unique local information encoding obtained for the coordinate $x^n$ by combining the features of the surrounding points. This encoding is only related to the number of sampling points selected based on distance and is independent of the order of the point sequence.

### C. The Forward Problem Calculation Method of EIT

The forward problem in EIT is used to calculate the boundary voltages given the conductivity distribution. The potential distribution function within the domain satisfies the Laplace equation with the conductivity distribution function:

$$\nabla \cdot [\sigma(x,y)\nabla\phi(x,y)] = 0, \quad (x,y) \in \Omega \quad (3)$$

Where $\Omega$ represents the domain of the field. Subsequently, the boundary potential values are calculated using the finite element method. The Laplace equation mentioned above can be expressed in matrix form as:

$$[A][\phi] = 0 \quad (4)$$

$$A_{ii} = \sum_{\substack{The\ cell\ e \\ containing\ vertax\ i}} C_{ii}^e, A_{ij} = \sum_{\substack{The\ cell\ e \\ containing\ edge\ ij}} C_{ij}^e \quad (5)$$

$$[C]_e = \frac{\varepsilon}{4\triangle}\begin{bmatrix} y_j-y_m & x_m-x_j \\ y_m-y_i & x_i-x_m \\ y_i-y_j & x_j-x_i \end{bmatrix}\begin{bmatrix} y_j-y_m & y_m-y_i & y_i-y_j \\ x_m-x_j & x_i-x_m & x_j-x_i \end{bmatrix} \quad (6)$$

Where $e$ is the index of the triangular element in the finite element method, $C$ is the element coefficient matrix, $\varepsilon$ is the conductivity of element $e$, $\triangle$ represents the area of the triangular element, and $x$ and $y$ are the coordinates of the three vertices of the triangular element. By applying the boundary conditions of current density to equation (4), the potential matrix $\phi$ for each node in the field can be calculated. Furthermore, the voltage can be obtained by measuring the potential difference between nodes.

## III. METHOD

MR - EIT is a dual - mode framework that can employ either a data - driven mode or an unsupervised learning mode based directly on voltage value loss iteration. As shown in Fig.1, the framework consists of an ordered feature extraction module and an unordered coordinate feature expression module. In the data - driven mode, both modules are used simultaneously, while in the unsupervised learning mode, only the unordered coordinate feature expression module is required.

### A. Ordered Feature Extraction Module

In the ordered feature extraction module of MR - EIT, the voltage data is represented as two - dimensional conductivity features through a neural network. The main purpose of designing the ordered feature extraction module in the data - driven image reconstruction mode is to utilize local operators without symmetric properties to perform a series of transformations and mappings on the voltage sequence. Inspired by LIIF, an auto - encoder - structured neural network that maps voltage data end - to - end to the two - dimensional pixel domain is used to query the initial features of each coordinate in the symmetric part of the framework. The neural network structure of the ordered feature extraction module employs both Transformer - Encoder and two - dimensional convolution as the encoder. The feature maps during the convolutional downsampling process are used to establish skip connections with the decoder, and bilinear interpolation is used for upsampling in the decoder. Finally, a feature map with a shape of (1,256,256) is obtained for subsequent sampling.

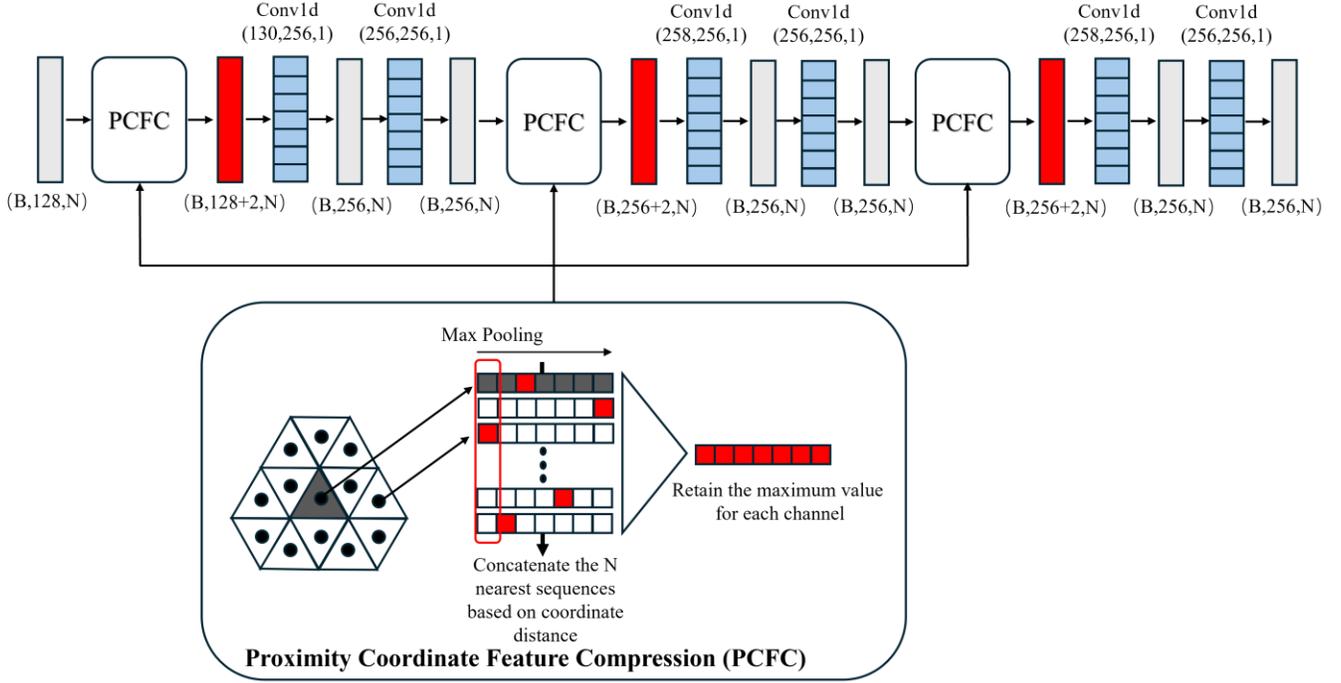

Fig. 3 Specific structure of the Feature Fusion Block.

For common triangular finite element meshes, the barycentric coordinates of each triangle are used to represent the entire element. When using data - driven methods, the value of the corresponding pixel in the pre - trained model's output feature map is queried using the coordinates. Since the coordinates of the two - dimensional feature map are discrete, the nearest neighbor sampling (nearest) method is used to obtain the feature values corresponding to the coordinates. Additionally, the number of finite element units in EIT is much smaller than the total number of pixels in the pre - trained model's output feature map. Therefore, the Feature Unfold method in equation (1) is used to expand the range of selected features. In the data - driven mode, the input to the symmetric module is $\{X_i, Y_i, \widehat{M}_{X_iY_i}\}$, that is, each coordinate is concatenated with the queried feature to form the input. Correspondingly, in the unsupervised learning mode, only $\{X_i, Y_i\}$ is used as the input to the symmetric module. As shown in Fig.2, in the data - driven mode, due to the additional coordinate attribute data being concatenated, the kernel size of the first one - dimensional convolution operation is 11. The rest of the calculation process is the same as in the unsupervised learning mode.

### B. Unordered Coordinate Feature Expression Module

The unordered coordinate feature expression module is the key to enabling MR - EIT to reconstruct images at multiple resolutions. Its computation is entirely independent of the order and quantity of the input data, ensuring efficient zero - sample transfer for image reconstruction under various resolutions or numbers of elements. Specifically, this module is mainly composed of a MLP constructed by one - dimensional convolution, the symmetric function in equation (2), and the proximity coordinate feature compression mechanism. As shown in Fig.2, except for the first one - dimensional convolutional layer, which differs due to the input dimension, the model design is exactly the same in both modes. To ensure that the same set of neural network parameters can be used for coordinate attribute feature vectors with different input orders and quantities, all computational mechanisms in this module are the same for each input feature vector and do not involve any local operator calculations that depend on the input order. All one - dimensional convolution operations use the length of the feature vector as the number of channels, with a constant kernel size of 1, and the length of the feature vector is changed by altering the number of channels. However, using only the MLP composed of one - dimensional convolution cannot capture the relationships between input feature vectors.

To obtain the local relationships that are crucial for image reconstruction, a local information extraction method composed of symmetric functions and a proximity coordinate feature compression mechanism is designed. The proximity element(coordinate) feature selection mechanism is used to select the N nearest coordinate points to a given point and concatenate the feature vectors of these N coordinate points. To maintain support for different finite element mesh element quantities, similarly, a computation method that is independent of the vector sequence order must be used to obtain the local information of these N feature vectors and represent this information as a one - dimensional vector. Then, as shown in Fig.3, max - pooling is used as a symmetric function to compress the features of these N vectors, retaining only the maximum value of each dimension among the N vectors. The N vectors are then re-compressed into a one-dimensional vector, which represents the local feature information of the corresponding coordinate point. Combining the symmetric function with the proximity coordinate feature compression mechanism can achieve a similar effect to two-dimensional convolution in extracting local information from a two-dimensional plane. Moreover, this approach does not require a specific sequence order for the input data. In this module, by continuously using local feature compression, the receptive field can be expanded. Since this computational mechanism only involves selecting neighboring coordinate vectors and max-pooling to compress vectors, and does not include any parameterized calculations, it can easily expand the local feature extraction range by modifying the number of neighboring elements selected in the data-driven mode for multi-resolution zero-sample transfer image reconstruction tasks without training additional parameters. In the iterative mode, it can provide the feature vectors of neighboring elements as supplements for each element, effectively enhancing the smoothness of the image and the distinction of boundaries.

### C. Data-driven Mode

The data-driven mode consists of two stages. The first stage involves pre-training the ordered feature extraction module, while the second stage trains both modules simultaneously. In the first stage, an auto-encoder is used to establish a feature mapping from voltage to the two-dimensional image domain. This auto-encoder is defined as the function:

$$M = f_\theta(v) \quad (7)$$

Under this definition, $v$ represents the voltage, and $M$ is the two-dimensional feature map. Each pixel in this feature map, which is defined in the pixel domain, represents a conductivity feature. To enable the unit centroid coordinates defined in the continuous two-dimensional coordinate domain to query and obtain features from the two-dimensional feature map defined in the pixel domain, the continuous coordinate domain of the unit centroid coordinates needs to be normalized:

$$(x', y') = \left(2 \cdot \frac{x - x_{min}}{x_{max} - x_{min}} - 1, 2 \cdot \frac{y - y_{min}}{y_{max} - y_{min}} - 1\right) \quad (8)$$

Where $(x', y')$ represents the normalized coordinates, and $(x, y)$ represents the original coordinates. Subsequently, it is easy to obtain the query coordinates corresponding to the pixel coordinate range through scaling or translation. Then, according to equation (1), the feature vector of the query point is obtained from the pixel-domain feature $M \in \mathbb{R}^{H \times W}$. The pre-training dataset consists of 20,000 pairs of voltage and corresponding two-dimensional pixel-domain conductivity data. Among them, 16,000 pairs are used as the training set, 2,000 pairs as the validation set, and 2,000 pairs as the test set. The loss function uses the mean squared error (MSE):

$$L_{pretrain} = \frac{1}{NHW} \sum_{i=1}^{N} \sum_{j=1}^{H} \sum_{k=1}^{W} (I_i(j,k) - M_i(j,k))^2 \quad (9)$$

Here, $I$ represents the pixel label values, $M$ represents the predicted values, and $(N, H, W)$ represent the batch size and the height and width of the two-dimensional image, respectively.

The second stage involves training the unordered coordinate feature expression module, which is defined as:

$$\hat{\epsilon} = p_\psi(x, y, \widehat{M}) \quad (10)$$

In this mode, the input to the unordered coordinate feature expression module $p_\psi$ includes the coordinates and the conductivity feature vector obtained through equation (2), and the output is the predicted element conductivity $\hat{\epsilon}$. The optimization objective for the neural network parameters is:

$$L_{train}(\theta, \psi) = argmin_{\theta,\psi} \frac{1}{N} \sum_{i=1}^{N} \|\epsilon - \hat{\epsilon}\|^2 \quad (11)$$

In the second stage, the parameters of both modules are trained simultaneously, where $\epsilon$ represents the true element conductivity. The dataset for this stage is the same as the voltage data used in the pre-training stage, with the conductivity data in one-dimensional form corresponding to the conductivity of each element. The simulated dataset is created using COMSOL and Matlab. To avoid the so-called "inverse crime," all simulated voltage data are generated using a finite element mesh with 5,696 elements and 2,927 nodes. To facilitate the verification of the model's zero-sample multi-resolution reconstruction performance, a low-resolution finite element mesh with 636 elements and 345 nodes is used during training to correspond to the element conductivity data associated with the voltage.

### D. Unsupervised Learning Mode

In this mode, unsupervised learning is employed, eliminating the need for training data. As shown in Fig.3, the input to the unordered coordinate feature expression module is the centroid coordinates of each mesh element. The symmetric function and the mechanism for selecting features of neighboring elements provide local information for each coordinate-conductivity encoding process and offer the functionality to store and update parameters in multi-mesh scenarios within the same field. Since the mechanism for obtaining local features does not involve trainable parameters, the number of parameters is low, allowing for excellent reconstruction results with fewer iterations. This mode also employs a two-stage execution process. For the field to be reconstructed, mesh divisions with both low and high numbers of elements are used. According to equation (6), a higher number of elements will require longer computation times. For the typical EIT inverse problem, the conductivity values themselves are usually treated as the parameters to be optimized:

$$\sigma = \underset{\sigma}{arg\min}\{\|V - U(\sigma)\|^2\} \quad (12)$$

Where $U$ is the function for computing the voltage in the EIT forward problem. When there is a need to upscale the number of reconstruction elements, $\sigma$ has to be reinitialized, and the optimized parameters from the previous iteration cannot be leveraged. However, The unordered coordinate feature expression module serves to store the outcomes of parameter updates within the neural network. For varying meshes within an identical field, the coordinates are fed into the neural network as inputs.

$$\hat{\epsilon} = p_\phi(x, y) \quad (13)$$

$$\phi = \underset{\phi}{arg\min}\{\|V - U(\hat{\epsilon})\|^2\} \quad (14)$$

The parameter optimization process utilizes the same neural network. Since all operators in the design of the unordered coordinate feature expression module are independent of the number and order of input data, the neural network parameters can be directly reused for the same field to be optimized. Overall, the two-stage execution mode adopted in this mode is as follows: The first stage uses a mesh with a lower number of elements to achieve reconstruction, aiming to make the neural network parameters $\phi$ converge in a shorter amount of time. In the experiments of this work, the average number of iterations for this stage is 200. The second stage employs a mesh with a significantly higher number of elements than the first stage for reconstruction, utilizing the neural network parameters $\phi$ from the first stage. In the experiments, the average number of iterations required is only around 50.

*E. Experimental Platform*

All experiments and training in this work were deployed on a Linux system. The neural network model was built using Pytorch 2.0, and a custom stiffness matrix parallel computing operator was constructed using CUDA. The hardware device used was an NVIDIA GEFORCE RTX 4070.

## IV. RESULT

MR-EIT is a method that can operate in both supervised and unsupervised learning modes. In this section, simulation experiments are designed to verify the image reconstruction performance of the method under supervised learning conditions and the reconstruction effect after changing the number of mesh elements. Since unsupervised learning has a broader range of applications than supervised learning, simulation and physical experiments are specifically designed to verify the unsupervised learning-based image reconstruction performance of MR-EIT and the image reconstruction effect after super-resolution.

*A. Evaluation Metrics*

Structural Similarity (SSIM) is used as an evaluation metric to measure the differences between the reconstructed image and the comparison image, as well as the structural similarity of the targets within the reconstructed image. Relative Image Error (RIE) is used to compare the differences between images. The definitions of RIE and SSIM are:

$$RIE = \frac{\|\hat{I} - I\|^2}{\|I\|^2} \quad (15)$$

$$SSIM(I, \hat{I}) = \frac{(2\mu_I \mu_{\hat{I}} + C_1)(2\sigma_{I\hat{I}} + C_2)}{(\mu_I^2 + \mu_{\hat{I}}^2 + C_1)(\sigma_I^2 + \sigma_{\hat{I}}^2 + C_2)} \quad (16)$$

where $\hat{I}$ represents the reconstructed image, I represents the real image data, $\mu_I$ represents the mean value of I, $\mu_{\hat{I}}$ represents the mean value of $\hat{I}$, $\sigma_I$ is the standard deviation of $I$, $\sigma_{\hat{I}}$ is the standard deviation of $\hat{I}$, $\sigma_{I\hat{I}}$ is the covariance of the two images of $I$ and $\hat{I}$, and $C_1$ and $C_2$ represent two constants.

*B. Performance Verification of the Data-driven Mode*

In this subsection, the data-driven mode of MR-EIT is compared with different types of image reconstruction algorithms. The first type of method uses neural networks constructed with sequential operators, including the most common U-Net structure and the more recent DiffusionEIT algorithm based on diffusion models. The significant characteristic of these methods is their strong learning ability for a certain deterministic sequential sequence. However, they cannot be directly transferred to a new sequence when the sequence changes. For example, in the case of the conductivity sequence in this paper, if the conductivity sequence is directly shuffled or a new sequence size is used, it is necessary to reconstruct the dataset and re-train from scratch. The second type of comparison method employs operators that are entirely sequence-independent, such as MLP or the Cross-Attention method in Transformer, combined with the unordered coordinate feature expression module. The main purpose of comparing these methods is to explore how to use voltage data as a condition for image reconstruction. In this paper, two common ways of incorporating voltage data into the coordinate-implicit neural representation module are mainly compared. One is to directly concatenate consistent voltage data for each coordinate, and the other is to use the cross-attention computation method. The hidden variables of the coordinates are used as Q, and the hidden variables of the voltage are used as K to calculate the attention scores. Then, the attention scores are decoded to obtain the final conductivity values corresponding to the coordinates.

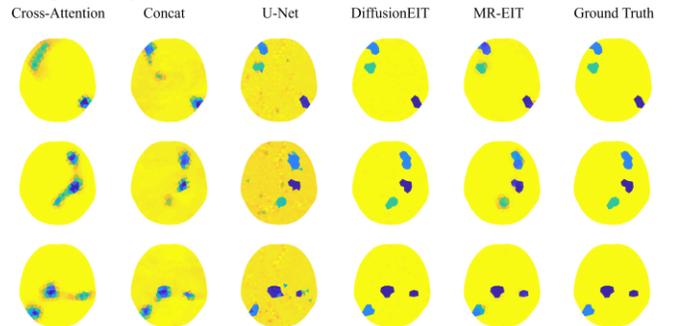

Fig. 4 Comparison of reconstruction results of MR-EIT using a data-driven model in the 636 elements condition.

Fig 4 shows the image reconstruction results at the resolution consistent with the training dataset. The image reconstruction effect using direct voltage concatenation is not ideal, especially in terms of image smoothness. Expressing each coordinate - voltage sequence is extremely challenging. This is because the global MLP cannot effectively capture the local correlations in the voltage data. Although the general location and target object values can be reconstructed, the loss of shape details is very severe. The method using the cross - attention mechanism to output attention scores has an incredible degree of image smoothness. This approach of directly using attention scores for subsequent calculations can regularize the differences in the output conductivity values to some extent. However, it fails to accurately identify the location and shape of the target objects.

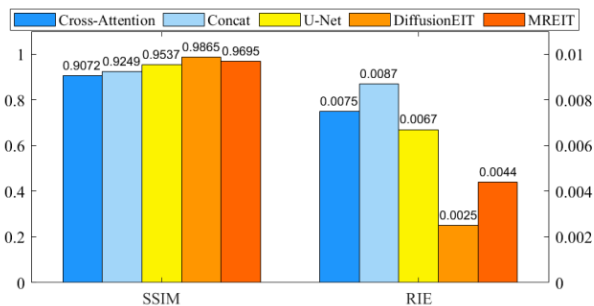

Fig. 5 The mean SSIM and RIE on the test set with the same number of elements as the training dataset.

Further comparison with methods based on sequential sequences shows that MR-EIT can achieve better image reconstruction than U-Net, but it is slightly inferior to the advanced sequential-based EIT image reconstruction method, DiffusionEIT. This result is not surprising. Diffusion models use operators that serve specific sequences and can easily learn local sequence information without considering the possibility of changes in the relative order of the sequence. Moreover, the image - generating capabilities of diffusion models have been widely recognized. Fig.5 shows the average SSIM and average RIE of the test set under the condition of 636 elements. The data-driven mode of MR-EIT can reflect the correct location and shape of the target objects and achieves better results compared to the coordinate-implicit expression methods based on concatenation and Cross-Attention.

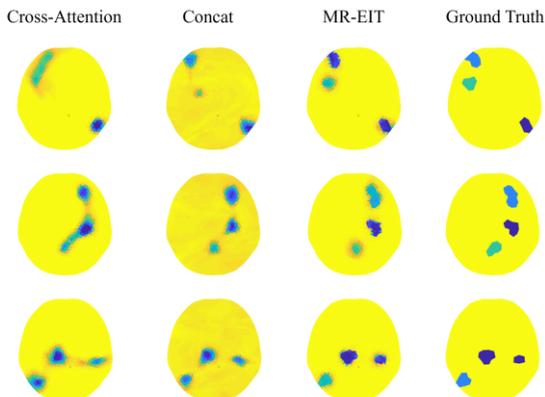

Fig. 6 The reconstruction performance comparison of MR-EIT in data - driven mode under the condition of 1752 elements.

Since neither U - Net nor DiffusionEIT has the capability for zero-shot multi-resolution image reconstruction when the number of mesh elements changes, the experiments with different numbers of elements only compared the coordinate-implicit expressions based on concatenation and Cross-Attention. The experimental design compared the reconstruction on 1,752 elements directly after training on 636 elements. As shown in Fig.6, MR-EIT can directly perform image reconstruction on a field with a higher number of mesh elements and the same shape without additional training samples, and it can almost maintain the original performance. The location and shape of the target objects do not show significant differences. In contrast, the methods based on concatenation and Cross-Attention fail to accurately reconstruct the targets. Fig.7 shows the performance metrics on the test dataset. MR - EIT can achieve relatively little performance loss when zero-shot transferring to image reconstruction with 1,752 elements. The comparison methods, due to insufficient learning of voltage representation, perform less ideally in image reconstruction with both the original and higher numbers of elements.

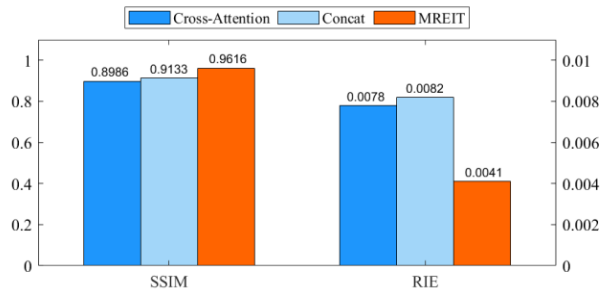

Fig. 7 The mean SSIM and RIE on the test set with a higher number of elements than the training dataset (1752 elements).

### C. Performance Verification of the Unsupervised Learning Mode

To verify the performance of the unsupervised learning mode of MR-EIT, simulation experiments and real - world tank experiments were designed. The unsupervised learning mode of MR-EIT does not require pre-training with a dataset, thus holding greater potential for application in physical experiments. The comparison methods selected were the widely - used L2 regularization method and the Gauss-Newton method. In the experiments, MR-EIT not only achieved accurate image reconstruction but also required fewer update iterations. On the hardware platform of this work, using parallel computing methods to calculate the stiffness matrix required for the EIT forward problem, the average time needed to update parameters once in image reconstruction with 636 elements was approximately 120 ms. In contrast, for image reconstruction with 5,696 elements, the average time required to update parameters once was about 1,500 ms. The total average number of iterations in the experiments generally did not exceed 300 epochs.

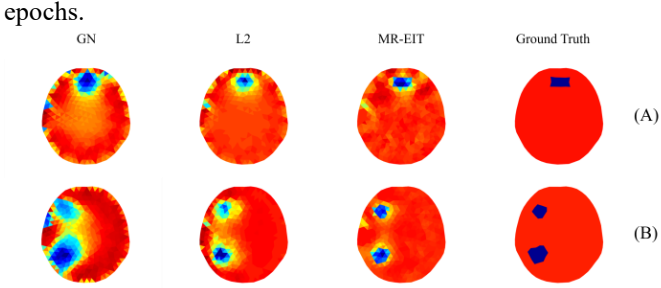

Fig. 8 The reconstruction performance of the unsupervised learning mode of MR-EIT on simulated samples.

The results of the simulation data are shown in Fig.8. MR-EIT achieves significantly better reconstruction results in terms of the location and shape of the target objects compared with the comparison methods. Moreover, there is a marked difference in the transition between the target objects and the background. TABLE I shows the quantitative comparison results of the evaluation metrics.

TABLE I

EVALUATION METRICS FOR SAMPLE (A) AND (B)

|  | (A) | | | (B) | | |
| --- | --- | --- | --- | --- | --- | --- |
|  | MR-EIT | GN | L2 | MR-EIT | GN | L2 |
| SSIM | 0.9114 | 0.9001 | 0.8603 | 0.9208 | 0.8822 | 0.8131 |
| RIE | 0.0580 | 0.1107 | 0.2145 | 0.0530 | 0.1217 | 0.3801 |

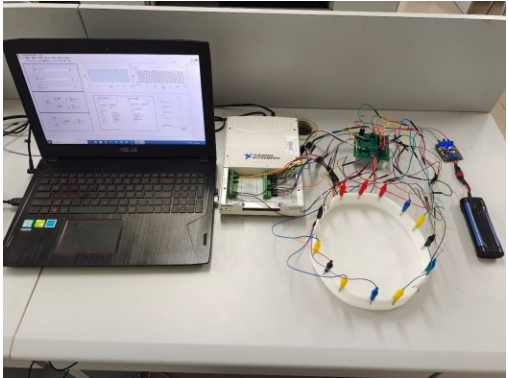

Fig. 9 Data acquisition device.

A real-world water tank experiment was designed to verify the performance of MR-EIT under actual conditions. The voltage measurement data acquisition system consists of the following components: an NI USB-6281 data acquisition card, a switch array composed of four 16-channel multiplexers (model ADG1606), a current source, wires, electrode clamps, and a personal computer. When the system operates, the current source converts the control voltage generated by the data acquisition card into an excitation current with a frequency of 1 kHz and an amplitude of 0.5 mA. The data acquisition card controls the switch array to select which electrodes to inject the excitation current into. Finally, the personal computer collects the voltage signals on the electrodes by controlling the NI USB-6281 data acquisition card. As shown in Fig.9, the measured field is a human skull-shaped tank that is approximately 22 cm long and 20 cm wide, using saline as the background conductive medium with a conductivity of about 1.5 S/m.

The main challenges faced in the image reconstruction of the actual experiment are the noise contained in the measured voltage, and the noise from the water surface fluctuation and electrodes is particularly severe during measurement. Therefore, the actual experiment can well reflect the sensitivity of the reconstruction algorithm to noise and its feasibility of application. The reconstruction strategy used by MR-EIT in this part of the experiment is the same as described in the methods section. It first uses the 636 elements reconstruction image with a fast iteration speed. Since the unordered coordinate feature expression module can handle coordinates of any number of elements and uses shared parameters, it has a faster neural network parameter iteration speed. Fig.10 shows the comparison of image reconstruction results on 636 elements. MR-EIT can accurately reconstruct the target objects, while the comparison method that directly takes the element conductivity values as parameters for optimization fails to reconstruct the correct basic shape. It is worth noting that in the case of 636 elements, equation (2) is applied to extract local information from the 16 nearest neighboring elements.

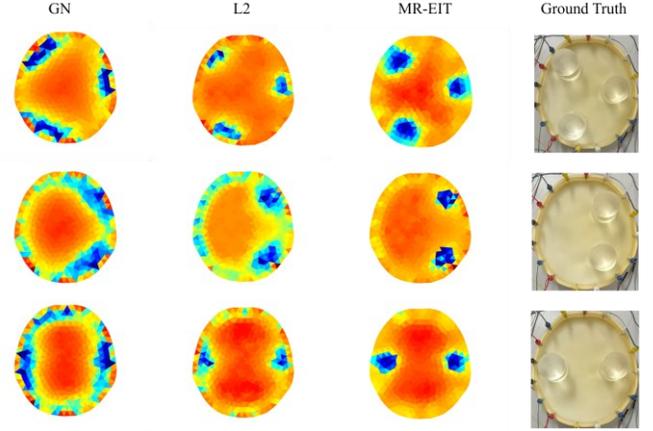

Fig. 10 The reconstruction results of the unsupervised learning mode of MR-EIT for real voltage samples on a grid with 636 elements (low resolution).

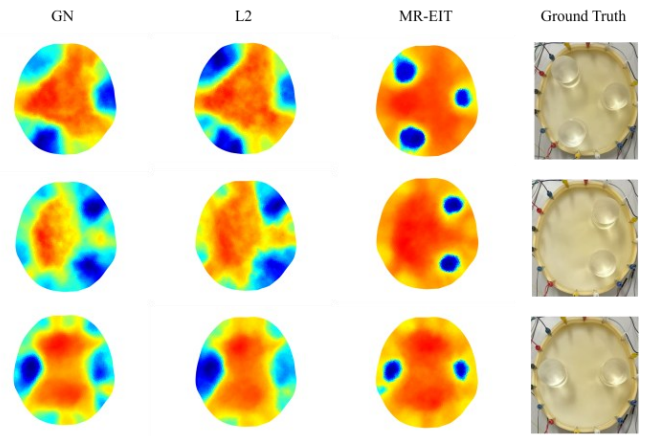

Fig. 11 The reconstruction results of the unsupervised learning mode of MR-EIT for real voltage samples on a grid with 5696 elements (high resolution).

Entering the super-resolution phase of the unsupervised mode, the previous stage has already completed image reconstruction for 636 elements, and the unordered coordinate feature expression module has been able to achieve good transformation of element coordinate data. At this point, the mesh is changed to 5,696 elements, and the number of extracted element information is correspondingly increased from 16 to 48. The parameter optimization is started again. Since the neural network parameters are shared, it only takes an average of fewer than 50 iterations to achieve image reconstruction with a higher number of elements. The higher resolution brings more delicate edge representation. In the actual experiment, more accurate shape expression can be achieved, and thanks to the increase in the number of elements, even without using any interpolation methods to enhance the image, the reconstruction effect is still relatively smooth. Fig.11 shows the conductivity element reconstruction results on 5,696 elements. Compared with the reconstruction image of 636 elements, the higher number of elements supplements the edge shape details missing due to the limitation of element shape, significantly improving the clarity and accuracy of the image.

## V. Conclusion

In this study, we proposed the MR - EIT method for multi-resolution reconstruction in Electrical Impedance Tomography (EIT). This method integrates both data-driven and unsupervised learning modes to achieve efficient and flexible image reconstruction at different resolutions. The ordered feature extraction module, with its hybrid architecture combining convolution and Transformer encoder, is capable of extracting both local and global features from voltage data. Meanwhile, the unordered coordinate feature expression module leverages symmetric functions and Proximity Coordinate Feature Compression mechanisms to realize multi-resolution image reconstruction. This dual - mode design enables MR- EIT to complete high quality image reconstruction using pre-trained models in the data-driven mode, while in the unsupervised mode, it can achieve multi-resolution image reconstruction without any dataset training.

Through comprehensive simulation experiments and real-world tank experiments, we verified the superior performance of MR-EIT in accurately reconstructing the location and shape of target objects, even under different mesh resolutions. The method exhibits strong robustness against noise and can generate smooth, clear images with minimal computational overhead. The ability to switch between data - driven and unsupervised modes makes MR-EIT adaptable to a variety of practical application scenarios.

## References

[1] B. Schullcke, B. Gong, S. Krueger-Ziolek, M. Soleimani, U. Mueller-Lisse, and K. Moeller, "Structural-functional lung imaging using a combined CT-EIT and a Discrete Cosine Transformation reconstruction method," Sci Rep, vol. 6, p. 25951, May 2016, doi: 10.1038/srep25951.
[2] S. Rana, R. Hampson and G. Dobie, "Breast Cancer: Model Reconstruction and Image Registration From Segmented Deformed Image Using Visual and Force Based Analysis," in IEEE Transactions on Medical Imaging, vol. 39, no. 5, pp. 1295-1305, May 2020, doi: 10.1109/TMI.2019.2946629.
[3] T. Dowrick, C. Blochet, and D. Holder, "In vivo bioimpedance changes during haemorrhagic and ischaemic stroke in rats: towards 3D stroke imaging using electrical impedance tomography," *Physiological Measurement*, vol. 37, no. 6, pp. 765–784, 2016.
[4] B. McDermott, A. Elahi, A. Santorelli, M. O'Halloran, J. Avery, and E. Porter, "Multi-frequency symmetry difference electrical impedancetomography with machine learning for human stroke diagnosis," Physiol.Meas., vol. 41, no. 7, Aug. 2020, Art. no. 075010.
[5] N. Goren et al., "Multi-frequency electrical impedance tomography and neuroimaging data in stroke patients," Sci. Data, vol. 5, no. 1, Jul. 2018,Art. no. 180112.
[6] B. Schullcke, B. Gong, S. Krueger-Ziolek, M. Soleimani, U. Mueller-Lisse, and K. Moeller, "Structural-functional lung imaging using a combined CT-EIT and a discrete cosine transformation reconstruction method," Sci. Rep., vol. 6, p. 25951, May 2016.
[7] V. A. Cherepenin *et al.*, "Three-dimensional EIT imaging of breast tissues: system design and clinical testing," *IEEE Transactions on Medical Imaging*, vol. 21, no. 6, pp. 662–667, 2002, doi: 10.1109/TMI.2002.800602.
[8] H. Zhang, Q. Wang, and N. Li, "DA-Net: A Dense Attention Reconstruction Network for Lung Electrical Impedance Tomography (EIT)," IEEE Internet of Things Journal, vol. 11, no. 12, pp. 22107–22115, Jun. 2024, doi: 10.1109/JIOT.2024.3380845.
[9] J. Huang et al., "EIT-MP: A 2-D Electrical Impedance Tomography Image Reconstruction Method Based on Mixed Precision Asymmetrical Neural Network for Hardware–Software Co-Optimization Platform," IEEE Sensors Journal, vol. 24, no. 23, pp. 39947–39957, Dec. 2024, doi: 10.1109/JSEN.2024.3476189.
[10] G. Ma, H. Chen, S. Dong, X. Wang, and S. Zhang, "PDCISTA-Net: Model-Driven Deep Learning Reconstruction Network for Electrical Impedance Tomography-Based Tactile Sensing," IEEE Transactions on Industrial Informatics, vol. 21, no. 1, pp. 633–642, Jan. 2025, doi: 10.1109/TII.2024.3456443.
[11] J. Sun et al., "Multi-Modal EIT Image Reconstruction Using Deep Similarity Prior," in 2024 IEEE International Instrumentation and Measurement Technology Conference (I2MTC), May 2024, pp. 1–6. doi: 10.1109/I2MTC60896.2024.10560635.
[12] K. Tang and C. Wang, "STSR-INR: Spatiotemporal super-resolution for multivariate time-varying volumetric data via implicit neural representation," Computers & Graphics, vol. 119, p. 103874, Apr. 2024, doi: 10.1016/j.cag.2024.01.001.
[13] Y. Han et al., "Super-NeRF: View-consistent Detail Generation for NeRF Super-resolution," IEEE Transactions on Visualization and Computer Graphics, pp. 1–14, 2024, doi: 10.1109/TVCG.2024.3490840.
[14] Y. Chen, S. Liu, and X. Wang, "Learning continuous image representation with local implicit image function," in Proceedings of the IEEE/CVF conference on computer vision and pattern recognition, 2021, pp. 8628–8638.
[15] O. Ronneberger, P. Fischer, and T. Brox, ''U-Net: Convolutional networks for biomedical image segmentation,'' Proc. Int. Conf. Med. Image Comput. Comput.-Assist. Intervent., 2015, pp. 234–241.
[16] J. Ho, A. Jain, and P. Abbeel, "Denoising diffusion probabilistic models," Proc. Adv. Neural Inf. Process. Syst. vol. 33, 2020, pp. 6840–6851.
[17] X. Han et al., "Semi-supervised model based on implicit neural representation and mutual learning (SIMN) for multi-center nasopharyngeal carcinoma segmentation on MRI," Computers in Biology and Medicine, vol. 175, p. 108368, Jun. 2024, doi: 10.1016/j.compbiomed.2024.108368.
[18] J. Feng et al., "Spatiotemporal implicit neural representation for unsupervised dynamic MRI reconstruction," IEEE Transactions on Medical Imaging, pp. 1–1, 2025, doi: 10.1109/TMI.2025.3526452.
[19] C. R. Qi, H. Su, K. Mo, and L. J. Guibas, "Pointnet: Deep learning on point sets for 3d classification and segmentation," in Proceedings of the IEEE conference on computer vision and pattern recognition, 2017, pp. 652–660.
[20] C. R. Qi, L. Yi, H. Su, and L. J. Guibas, "Pointnet++: Deep hierarchical feature learning on point sets in a metric space," Advances in neural information processing systems, vol. 30, 2017.
[21] G. Qian et al., "Pointnext: Revisiting pointnet++ with improved training and scaling strategies," Advances in neural information processing systems, vol. 35, pp. 23192–23204, 2022.